\documentclass[a4paper]{esannV2}
\usepackage[dvips]{graphicx}
\usepackage{amssymb,amsmath,array}
\usepackage{pgfplots}
\usepackage{float}
\usepackage{hyperref}
\usepackage{pdfpages}
\usepackage{subcaption}
\usepackage{wrapfig}

\newcommand{\mathdefault}[1][]{}

\begin{document}
\title{Robustness and Regularization in Hierarchical Re-Basin}

\author{Benedikt Franke, Florian Heinrich, Markus Lange \\
and Arne Raulf
%
%
\vspace{.3cm}\\
%
German Aerospace Center (DLR) - Institute for AI Safety and Security \\
Wilhelm-Runge-Strasse 10, 89081 Ulm - Germany \\
%
}

\maketitle
\vspace{-4em}
\begin{figure}[h!]
     \centering
     \begin{subfigure}[b]{0.15\textwidth}
         \begin{center}
            \includegraphics[width=\textwidth]{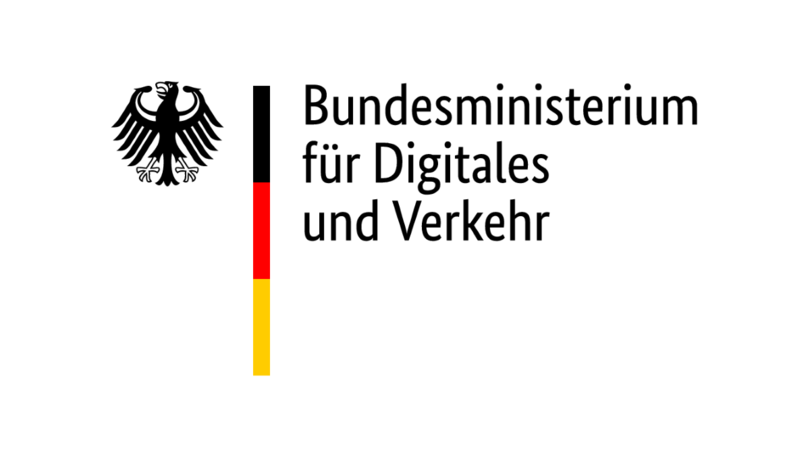}
         \end{center}
     \end{subfigure}
     \begin{subfigure}[b]{0.15\textwidth}
         \begin{center}
            \includegraphics[width=\textwidth]{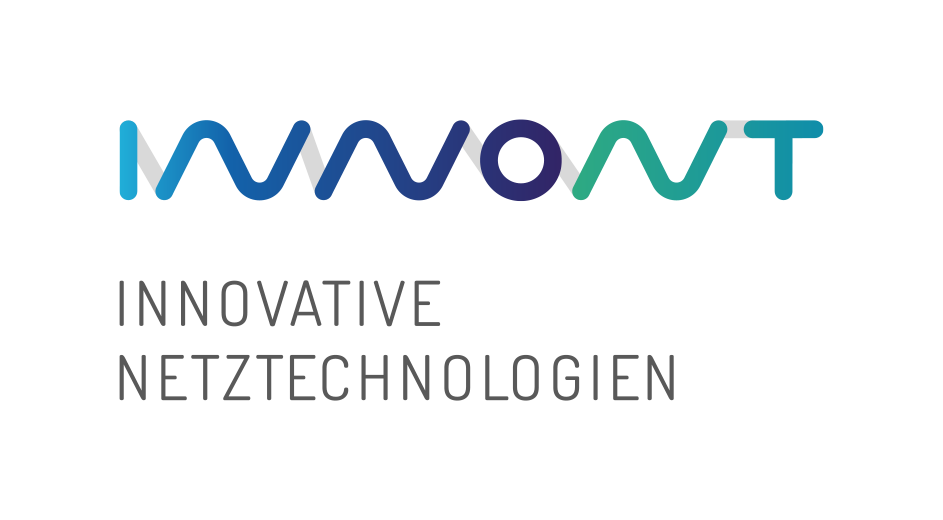}
         \end{center}
     \end{subfigure}
     \label{fig:histograms}
\end{figure}

\begin{abstract}
This paper takes a closer look at Git Re-Basin, an interesting new approach to merge trained models. 
We propose a hierarchical model merging scheme that significantly outperforms the standard \textit{MergeMany} algorithm.
With our new algorithm, we find that Re-Basin induces adversarial and perturbation robustness into the merged models, with the effect becoming stronger the more models participate in the hierarchical merging scheme.
However, in our experiments Re-Basin induces a much bigger performance drop than reported by the original authors.
\end{abstract}    


\section{Introduction}
\label{sec:intro}

In recent years, a lot of work has been done towards investigating the permutation symmetries of artificial neural networks (ANNs) trained with variants of stochastic gradient descent (SGD) \cite{frankle2020linear, conf/iclr/FrankleC19,mirzadeh2020linear, akash2022wasserstein, ainsworth2022git}. 
Hypothesized to be closely related to linear mode connectivity (LMC), permutation invariance allows to change the order of neurons in an ANN without hurting its accuracy \cite{DBLP:conf/iclr/EntezariSSN22}.
As a direct application of this hypothesis, Git Re-Basin \cite{ainsworth2022git} provides a method to merge two models with the same architecture by exploiting permutation invariance to "teleport" one of the models into the same loss basin as the other. 
In contrast to simple, "naive" interpolation between two models, this technique circumvents accuracy losses that arise when the mean of both models in the parameter space falls outside of a loss basin or local minima.
Motivated by these results, we investigate how to effectively merge more than two models.
We propose a hierarchical scheme as an alternative to the \textit{MergeMany} algorithm provided by \cite{ainsworth2022git} 
and show that it performs significantly better, even though it likewise does not admit a zero-accuracy barrier.
We also show that Git Re-Basin seems to induce adversarial robustness properties as well as regularization, which grow more pronounced the more models are merged with our scheme.
This effect already starts to be noticeable with just two input models.


\section{Related Work}
\label{sec:related}
LMC as first hypothesized in \cite{frankle2020linear} describes the property of SGD solutions to be linearly connected in the loss landscape.
While originally investigated in the context of the lottery ticket hypothesis \cite{conf/iclr/FrankleC19}, LMC has quickly become of interest to other sub-fields, such as multitask learning \cite{mirzadeh2020linear}.
Akash et al.  propose a mathematical framework that can use LMC towards model fusion \cite{akash2022wasserstein}.
General mode connectivity for language models is studied in \cite{qin2022exploring}.
Recently, LMC has been generalized to layerwise linear feature connectivity \cite{zhou2024going}, where it is claimed that not only the weight vectors of entire networks, but also single feature maps have a linear connection between differently trained networks.
Model Permutation has been linked to LMC \cite{DBLP:conf/iclr/EntezariSSN22}, which proposes LMC to only hold when taking into account \textit{permutation invariance}, i.e. that there exist permutations of trained networks which yield the same outputs as their originals.
Permutation invariance and its application towards model interpolation is studied in \cite{ainsworth2022git}.
In the context of federated machine learning, Wang et al. show that accuracy can be improved by permuting one of the models so that it closely matches the other weight-wise before averaging \cite{wang2020federated}.
Permutation invariance has also been studied in the context of adversarial attacks on ANNs \cite{ganju2018property}.

\section{Our Method}
\label{sec:method}
\begin{wrapfigure}{r}{0.5\textwidth}
    \includegraphics[width=0.5\columnwidth]{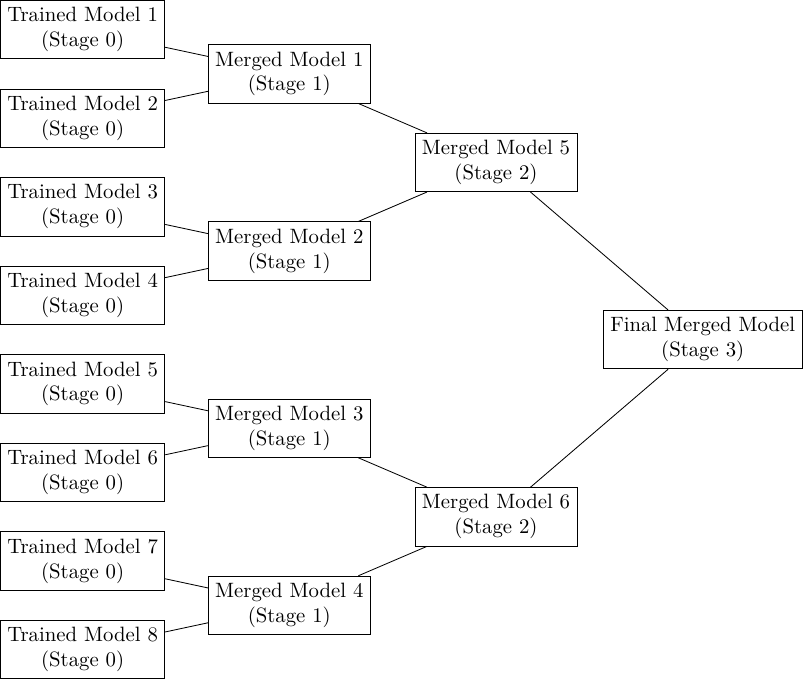}
    \caption{Our proposed hierarchical Merging Scheme, exemplified for merging eight models.}
    \label{fig:merging-scheme}
\end{wrapfigure}
While \cite{ainsworth2022git} provides the \textit{MergeMany} algorithm to apply Git Re-Basin to more than 2 models, we found the algorithm to have an important theoretical weakness:

In each round of the algorithm, one of the $n$ input models $\Theta_i$ is permuted towards the \textit{mean} $\bar{\Theta}$ of the other $n-1$ models with $\bar{\Theta} = \frac{1}{n} \sum_{j \in \{1,\ldots n\} \setminus {i}} \Theta_j$ \cite{ainsworth2022git}. 
However, the mean of these models can not be guaranteed to lie within a loss basin, despite the Re-Basin algorithm assuming all input models to lie within such a basin.
On the contrary, \cite{ainsworth2022git} shows that in general, linear interpolation between two non-permuted models yields strongly deteriorated accuracy. In Section \ref{sec:experiments}, we show that these theoretical considerations are also empirically observable. \\
To alleviate the discussed issue, we propose to use a more computationally expensive hierarchical scheme, which is outlined in Figure \ref{fig:merging-scheme}.
When trying to merge $2^n$ models, we merge them pairwise (and disjunctively) over $n$ stages, with each stage taking in the merged models of the previous stages.
In the following, we refer as "merging" to the process of applying Re-Basin to permute the second of two models to "teleport" it into the loss basin of the first model, and then doing linear interpolation as specified in \cite{ainsworth2022git} with $\lambda = 0.5$. 
In preliminary experiments, we investigated if it matters which model is chosen as the model that gets permuted, but found it to have no statistically significant influence.

\section{Experiments}
\label{sec:experiments}
\vspace{-1.7em}
\begin{figure}[h!]
     \centering
     \begin{subfigure}[b]{0.48\textwidth}
         \begin{center}
            \includegraphics[width=\textwidth]{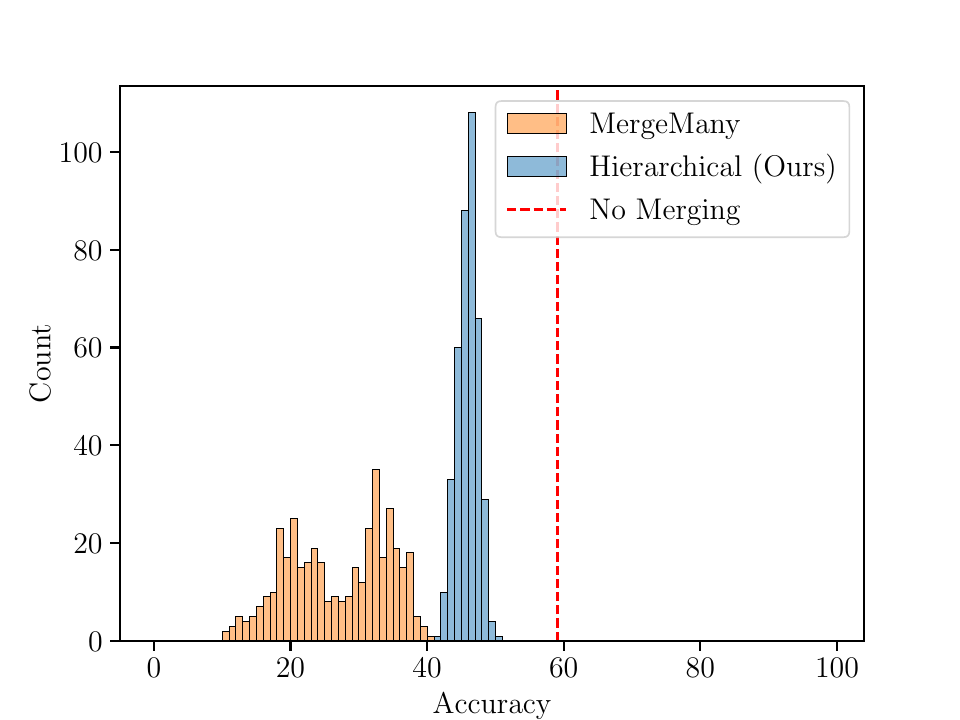}
         \end{center}
         \caption{Stage 2 (Merging 4 Models into one)}
         \label{fig:histograms:stage2}
     \end{subfigure}
     \hfill
     \begin{subfigure}[b]{0.48\textwidth}
         \begin{center}
            \includegraphics[width=\textwidth]{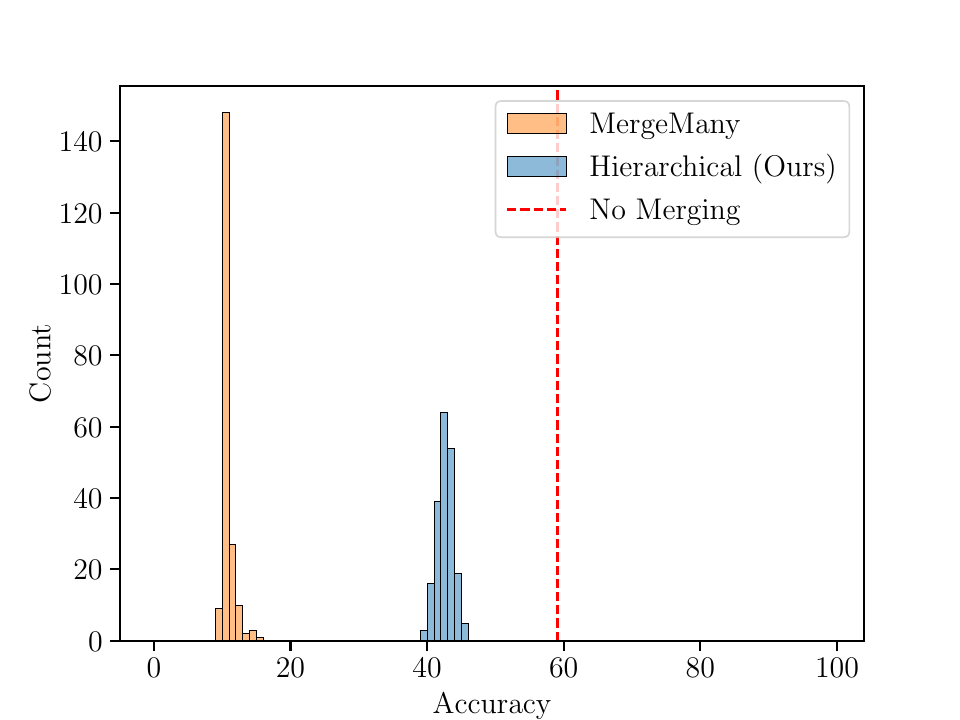}
         \end{center}
         \caption{Stage 3 (Merging 8 Models into one)}
         \label{fig:histograms:stage3}
     \end{subfigure}
     \caption{Distribution of test set accuracies on CIFAR-10 by merging algorithm on 1600 trained input models. "No Merging" denotes the median accuracy of the trained input models and is displayed only as a line as the unmerged models have very little variance to improve the clarity of the figure.}
     \label{fig:histograms}
\end{figure}

In order to investigate the effects of our approach in a statistically sound way, we train 1600 multi-layer perceptrons (MLPs) on the CIFAR-10 
data set.
We follow the experiments of \cite{ainsworth2022git} so that our MLPs have 4 layers with a hidden dimensions of 512 and a latent layer, which has a hidden dimension of 256. 
All layers except the last use the $ReLU$-activation and we use a different random seed for each training process.
Our experimental code is adapted from a PyTorch-based re-implementation of Re-Basin \footnote{Code taken from \url{https://github.com/themrzmaster/git-re-basin-pytorch}  - The Repository is not affiliated with the authors of this paper!}.
As a first test, we compare if our hierarchical Re-Basin yields better accuracy scores than the \textit{MergeMany} algorithm, by merging our 1600 trained models disjunctively into 400 models (Figure \ref{fig:histograms:stage2}) and 200 models (Figure \ref{fig:histograms:stage3}), respectively.
From the figures, we see that not only our Hierarchical Re-Basin approach outperforms the \textit{MergeMany} algorithm, but the difference gets stronger with more models merged.
This stems from the strong deterioration of accuracy in \textit{MergeMany} when switching from 4 input models to 8 input models.
On the contrary, our hierarchical approach does not suffer from the same effect, with only the variance of scores getting slightly larger.

\subsection{Robustness and Relationship with Regularization}
\begin{figure}[h!]
     \centering
     \begin{subfigure}[b]{0.48\textwidth}
         \begin{center}
            \includegraphics[width=\textwidth]{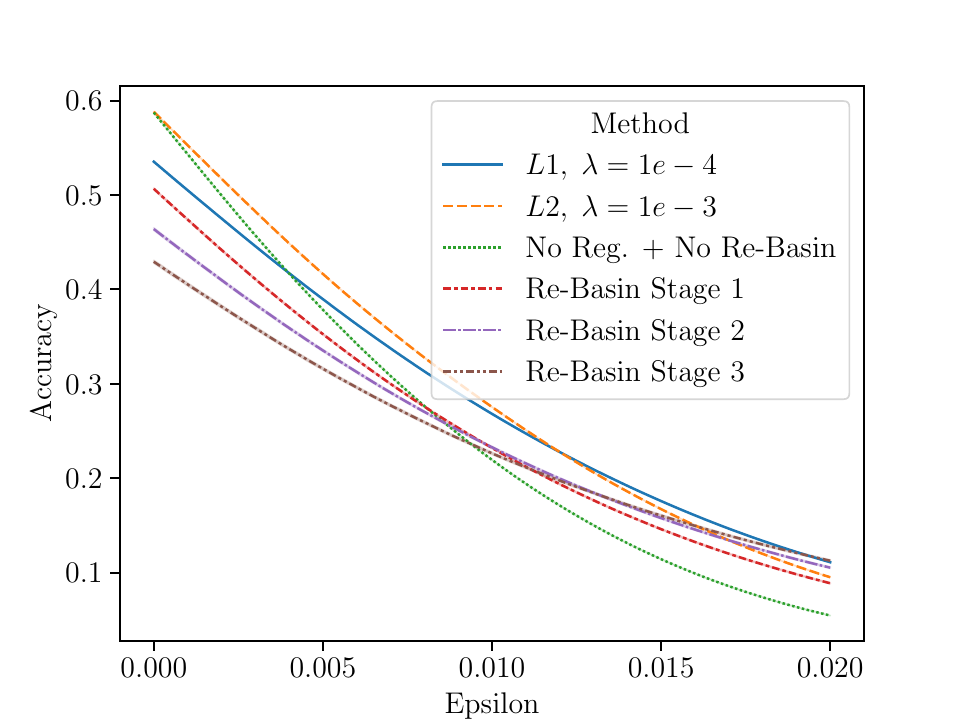}
         \end{center}
         \caption{DeepFool}
         \label{fig:adversarial:deepfool}
     \end{subfigure}
     \hfill
     \begin{subfigure}[b]{0.48\textwidth}
         \begin{center}
            \includegraphics[width=\textwidth]{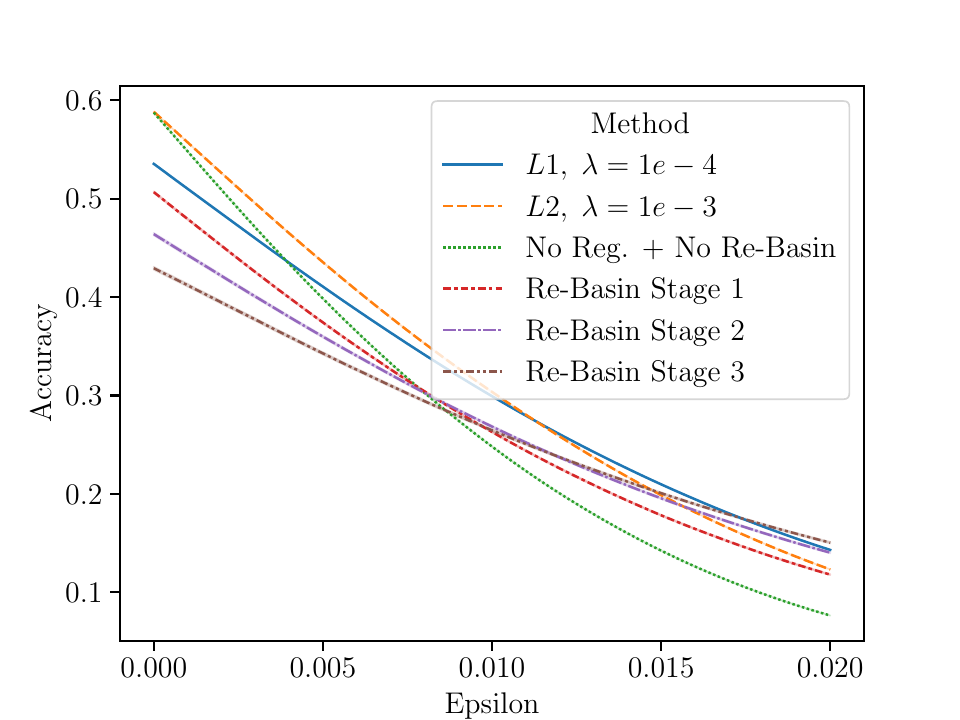}
         \end{center}
         \caption{FastGradient}
         \label{fig:adversarial:FGSM}
     \end{subfigure}
     \caption{Mean Accuracy of different Re-Basin stages over attack strength $\epsilon$. As a comparison, models trained with L1 and L2 regularization are also shown. 95\% CI is shaded around the curves but is too small to be visible.}
     \label{fig:adversarial}
\end{figure}

We further investigate if the merging of models via Re-Basin produces safety properties by testing our models for adversarial robustness on two selected attacks, DeepFool \cite{deepfool} and FGSM \cite{fgsm}.
The results are shown in Figure \ref{fig:adversarial}.
In both cases, we plot the robust accuracy (i.e. the accuracy on adversarial images) over the $\epsilon$-parameter, which controls the attack strength.
Figure \ref{fig:adversarial} shows that in both cases, all stages of Re-Basin break even with the unmerged models in the area around $\epsilon = 0.01$, with lower stages (i.e. less Re-Basin) being better with weaker attacks and later stages (i.e. more Re-Basin) being more robust against stronger attacks.
Overall, merging more models with Re-Basin seems to correlate with stronger adversarial robustness, but comes at the cost of some accuracy.
We also conducted the same experiments with L1 and L2 regularized models and plot the results similarly into Figure \ref{fig:adversarial}.
From the results, we see that L2 regularization seems to generally produce stronger adversarial robustness, only breaking even with the final stage of our multi-stage Re-Basin approach at the far end of the investigated $\epsilon$-range.

\subsection{Effects on Model Norm and Lipschitz Constant}
\begin{figure}[h!]
     \centering
     \begin{subfigure}[t]{0.48\textwidth}
         \begin{center}
            \includegraphics[width=\textwidth]{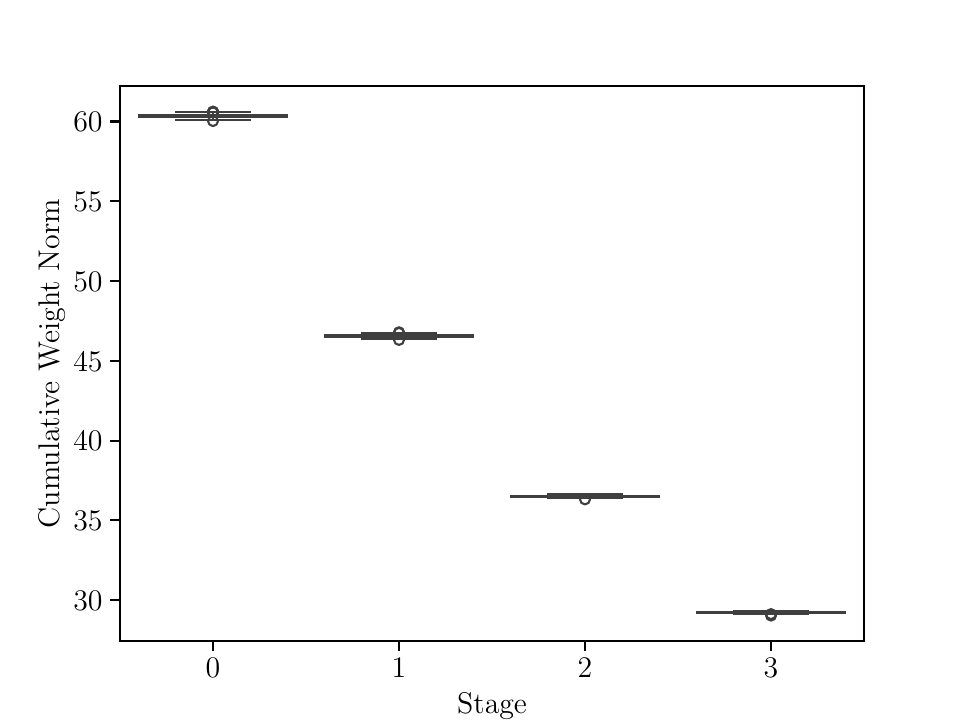}
         \end{center}
         \caption{Summed Frobenius Norm over Layers of MLP}
         \label{fig:norm}
     \end{subfigure}
     \hfill
     \begin{subfigure}[t]{0.48\textwidth}
         \begin{center}
            \includegraphics[width=\textwidth]{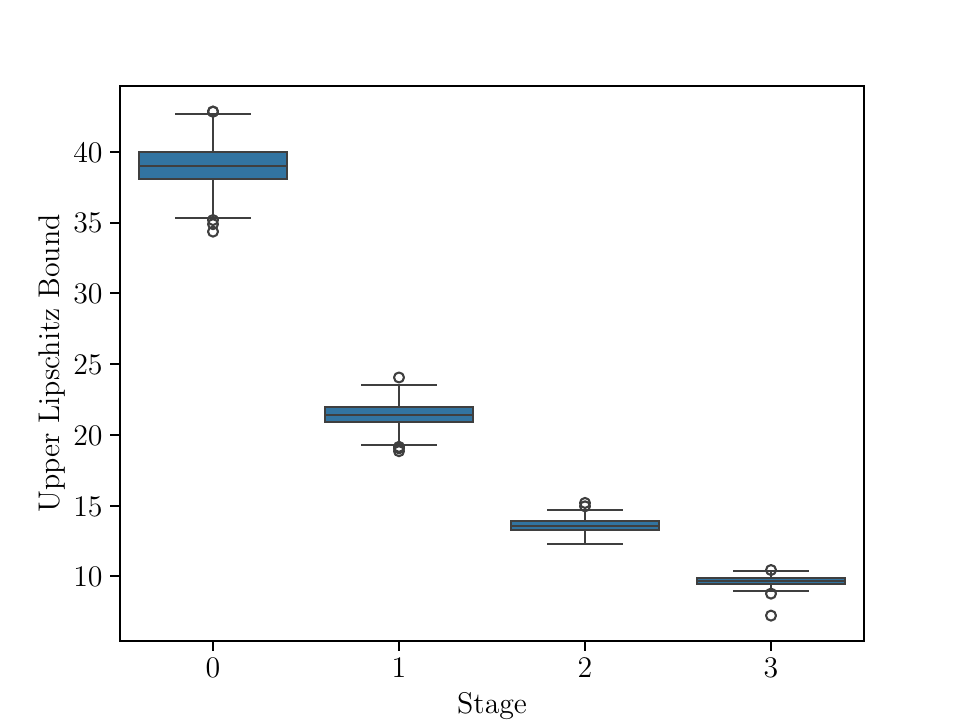}
         \end{center}
         \caption{Lipschitz Upper Bound}
         \label{fig:lipschitz}
     \end{subfigure}
     \caption{Impact of Different Re-Basin Stages on Weight Norm and Lipschitz Upper Bound}
\end{figure}

With the results on adversarial robustness in mind, we wonder what other properties of the MLPs are positively impacted by Re-Basin.
Firstly, we expect that Re-Basin should induce some weight-regularization into the resulting network, as noise-like perturbations on the weights are well-known to encourage smoother learned functions \cite{noise}. 
Interpolation between the weights of two trained networks can be expected to have a noise-like effect.
We use a very simple formula to calculate a metric for the weight regularization, $|w| = \sum_{i = 0}^{N} ||W_i||_F + ||b_i||_2$, where $N$ is the number of layers in the MLP, $W_i$ is the weight matrix of the $i$-th layer and $b_i$ is the bias of the $i$-th layer.
Intuitively, this measure produces higher values for less regularized models, as weight regularization commonly punishes high norms in the models parameters (e.g. L1 or L2 regularization \cite{ng2004feature}).
The distributions of this cumulative weight norm $|w|$ are depicted per stage of our Hierarchical Re-Basin in Figure \ref{fig:norm}.
Our findings support our initial expectation: $w$ drops almost linearly with the stage, while the variance also gets reduced.
Secondly, we investigate a well-known safety property of ANNs, the Lipschitz constant.
Employing the methodology\footnote{Code taken from \url{https://github.com/HeinrichAD/local_lipschitz/tree/778d7dec56574e43bdff05e8b6e794cb785c9d21}} presented in \cite{avant2023analytical}, we calculate a guaranteed upper bound on the Lipschitz constant of each MLP.
This upper bound is an effective measure for the sensitivity of an ANN to small input perturbations, which implies that a more robust network will have a smaller bound.
Our results are presented in Figure \ref{fig:lipschitz}, where a similar trend to Figure \ref{fig:norm} is clearly visible.
With each stage of Re-Basin, not only does the upper bound on the Lipschitz constant get smaller, but the variance in the bounds between the MLPs shrinks similarly. 
As the Lipschitz constant has been linked to perturbation resistance \cite{avant2023analytical}, this indicates a higher perturbation resistance induced by Re-Basin.

\section{Discussion}
\label{sec:conclusion}
In this paper, we presented a Hierarchical Re-Basin scheme and investigated the effects of Re-Basin on adversarial robustness, weight regularization and the Lipschitz constant.
We find that Re-Basin seems to act as a sort of regularization, positively impacting adversarial and perturbation robustness in the process.
This seems to stem from the interpolation step between two models in the same loss basin, which can be seen as a limited disturbance on the model weights that, by definition, should not move the model to a spot outside of the loss basin.
While any sufficiently strong noise on the model weights or data can act as such a regularization, we find that Re-Basin does so without hurting accuracy too much when applied using our proposed hierarchical Re-Basin scheme.
On the other hand, using the \textit{MergeMany} algorithm or even naively interpolating between models hurts accuracy so much that any emergent safety-properties become more or less irrelevant.
Our findings contradict \cite{ainsworth2022git}, which saw good results on their \textit{MergeMany} algorithm, despite the theoretical flaws discussed above.
Furthermore, despite using an open-sourced implementation of Re-Basin that was tested by multiple unaffiliated contributors, we were unable to reproduce the zero-accuracy barrier that was shown in the original paper.
We hope that the empirical results of this work contribute towards the ongoing discussion around Git Re-Basin and LMC and serve as a stepping stone for further investigations.



\begin{footnotesize}


\bibliographystyle{unsrt}
\bibliography{main}

\end{footnotesize}


\end{document}